\documentclass{article}

\usepackage[final]{neurips_2019}

\usepackage[utf8]{inputenc} 
\usepackage[T1]{fontenc}    
\usepackage{hyperref}       
\usepackage{url}            
\usepackage{booktabs}       
\usepackage{amsfonts}       
\usepackage{nicefrac}       
\usepackage{microtype}      
\title{AI Ethics for Systemic Issues: A Structural Approach}

\author{%
  Agnes Schim van der Loeff\thanks{Work completed as part of the Cervest Research Residency program. Correspondence to \texttt{\{agnes, jev\}@cervest.earth}} \qquad Iggy Bassi \qquad Sachin Kapila \qquad Jevgenij Gamper\thanks{Mentor.} \\
  \\
   Cervest Ltd. \\
   London, UK
   }

\begin{document}

\maketitle 

\begin{abstract}
The debate on AI ethics largely focuses on technical improvements and stronger regulation to prevent accidents or misuse of AI, with solutions relying on holding individual actors accountable for responsible AI development. While useful and necessary, we argue that this "agency" approach disregards more indirect and complex risks resulting from AI's interaction with the socio-economic and political context. This paper calls for a "structural" approach to assessing AI's effects in order to understand and prevent such systemic risks where no individual can be held accountable for the broader negative impacts. This is particularly relevant for AI applied to systemic issues such as climate change and food security which require political solutions and global cooperation. To properly address the wide range of AI risks and ensure 'AI for social good', agency-focused policies must be complemented by policies informed by a structural approach.
\end{abstract}

\section{Introduction}
\label{introduction}

In recent years it has become clear that while AI can greatly benefit society, it also carries many risks, exemplified by the racial bias in Google’s search engine, the use of facial recognition technology for surveillance, and Cambridge Analytica’s use of AI for targeted political messaging. This has resulted in growing calls for 'ethical AI' from critics while AI organisations each present their own set of ethical principles, the most common ones being transparency, fairness, non-maleficence, responsibility, and privacy \cite{jobin_artificial_2019}. Additionally, several initiatives have emerged to address AI ethics at a larger scale, such as the Montreal Declaration and the EU's guidelines for trustworthy AI. The general focus is on technical improvements in design and distribution on the one hand and better regulation on the other, where individual agents can be identified and held accountable for the ethical development, deployment and use of AI. While these are necessary steps in making AI more ethical, this paper questions whether they are sufficient to avoid the wide range of AI's potentially negative impacts. We argue that the dominant discourse neglects systemic risks, and needs to be complemented by a structural approach to account for the more indirect and complex causes of AI risks. Since these cannot be solved by technical changes alone, this paper calls for more interdisciplinary research and urges the AI community to engage with social scientists and policy makers. The first section explores to what extent the current approach, focused on individual agency, helps to reduce AI’s negative impacts. The second section assesses the limitations of this approach by contrasting it with a structural approach adopting a contextual understanding of AI’s systemic risks. The third section explores such risks in relation to climate change and food security, where AI’s interaction with socio-economic and political factors could have far-reaching impacts. Finally the paper offers some preliminary suggestions on how a structural approach might be implemented to help mitigate AI's harmful effects and ensure it is socially beneficial.

\section{Agency approach: technical improvements and better regulation}
\label{current approach}

The dominant discourse on AI ethics explores how negative outcomes can be prevented by improving the algorithmic design and optimising the distribution of technology. In an evaluation of AI ethics guidelines, the most often mentioned issues were those that could be solved to some extent by ‘technical fixes’ such as fairness, explainability, privacy, transparency and robustness \cite{hagendorff_ethics_2019}. For example, Google adjusted its racially biased algorithm, but without more fundamental changes to preventing bias in general; although this is a common response to ethical issues in the AI industry, such a solution is ‘ad hoc and reactive’ rather than a transformative shift to being more ethical \cite{crawford_there_2016}. Nonetheless, technical improvements are crucial to minimising certain risks while maximising benefits and thus optimising AI systems. This optimisation pertains to the production as well as distribution phase since negative impacts may not result from the product itself but its use and malicious use. Accordingly, in addition to ensuring the design is safe and secure, AI developers should carefully assess who their product, data or knowledge is shared with and to what extent. Optimisation of facial recognition technology could then entail not sharing it with actors that may use it maliciously, just as Amazon is being pressured to stop selling its ‘Rekognition’ system to government agencies \cite{whittaker_et_al_ai_2018}. In extreme cases such as automatic weapons it might entail halting development completely, although the current literature precludes this possibility, being concerned with making AI ‘better’ through technical improvements \cite{greene_better_2019}.

A second and equally vital emphasis in AI ethics literature has been on holding those responsible for AI’s negative impacts accountable. This would require new policies and enforcement mechanisms since existing frameworks governing AI do not successfully ensure accountability, being largely left to corporate self-governance \cite{whittaker_et_al_ai_2018}. Moreover, competition and lack of regulation mean that currently fast development is prioritised over ‘safe, secure and socially beneficial’ development \cite{askell_role_2019}. Regulation and liability laws are therefore needed to increase the incentive for responsible AI development \cite{askell_role_2019}. Otherwise, the plethora of ethics guidelines have no substantial impact and may further postpone legally binding regulations while their generic nature encourages ‘the devolution of ethical responsibility to others’ \cite{hagendorff_ethics_2019}. Accountability means that Google would face legal ramifications for releasing a biased algorithm, as would AI companies selling facial recognition technology to actors using it for unethical purposes, while those actors themselves would also be held accountable. For instance, the UK privacy regulator is currently assessing a private company’s use of facial recognition in CCTV systems in central London after its legality was questioned \cite{sabbagh_regulator_2019}. However, such AI governance might not be sufficient. The Cambridge Analytica scandal centred on the illegal sharing and using of data, not the use of AI for political messaging. Because public and regulatory energy focused on data sharing rather than the structural way AI can shape democratic processes by facilitating such ‘targeted propaganda’ \cite{brundage_et_al_malicious_2018}, it does not preclude widespread use of AI for political micro-targeting. No individual agents could be held accountable for these broader negative impacts, because the causes are structural rather than agency-based. 

\section{Structural approach: understanding systemic risks} \label{systemic risks}

The dominant discourse on AI ethics takes an agency approach in the sense that it addresses safety and security issues for which specific actors are responsible. However, it largely neglects systemic risks, which do not stem from either accidents or malicious use \cite{dafoe_ai_2018}. Instead, they result from the way AI systems shape and are shaped by the social, economic and political environment \cite{zwetsloot_thinking_2019}. A structural approach is needed to understand these effects which are more complex and indirect, and cannot be traced to individual actions \cite{zwetsloot_thinking_2019}. These characteristics are arguably a contributing factor to the general disregard for systemic risks, since obvious benefits outweigh opaque harmful effects, while their long-term nature is not easily captured by either corporate or political 'short-termism' \cite{paulson_jr._short-termism_2015}. Moreover, the distributed responsibility for these risks complicates the creation of necessary regulation and its enforcement \cite{ploug_should_2018}. Despite and because of this, a structural approach is needed to understand systemic risks. Lessons should be drawn from other fields, such as systemic risks in the financial sector and medical research. For instance, reviewers of a study indicating that antibiotics helped alleviate symptoms of a certain disease expressed concern at publishing it because despite the benefits, publication could inadvertently lead to a higher intake of antibiotics and thus increase antibiotic resistance \cite{ploug_should_2018}. Without any agent behaving unethically, the research’s unintended consequences could have negative societal impacts. Systemic risks thus apply to other industries, however it has particular relevance to AI because of the scale on which AI systems can operate and how rapidly they are evolving, with their impact growing as automation increases. Moreover, while ethics codes are starting to resemble the four principles of medical ethics, this is unlikely to translate into equally ethical \textit{practices}. Key differences between the two domains mean that medicine's 'principled approach' is unlikely to be as successful for AI, not in the least because AI lacks a clear common aim while the large share of private AI development allows commercial interests to trump public interests \cite{mittelstadt_ai_2019}. In AI ethics literature, the few times systemic risks are acknowledged, it is in relation to warfare (eg \cite{dafoe_ai_2018}) or competition in the ‘AI race’ to get first-mover advantage (eg \cite{hagendorff_ethics_2019}, \cite{cave_ai_2018}, \cite{askell_role_2019}). Race dynamics between companies, but especially between countries, compromise safety measures and ethical standards \cite{cave_ai_2018}. The race is therefore itself a systemic risk, caused by AI’s interaction with global politics and a competitive market economy lacking sufficient governance for AI ethics \cite{askell_role_2019}. To properly address this and other systemic risks posed by AI, agency-focused policies must therefore be complemented by policies informed by a structural approach.

\section{AI for climate change and food security}

One domain where a structural perspective is particularly crucial is climate change, because it is in itself a systemic issue linked to international politics and the global economic system. It is deeply political because responsibility and impact are unevenly distributed, since low income countries have generally not shared equally in the benefits of fossil fuels but are still harmed by the high income countries’ energy consumption \cite{diffenbaugh_global_2019}. Moreover, there is a strong inverse relationship between local impact of climate change and that location’s wealth such that ‘the greatest shifts in climate will be experienced by the poorest’ \cite{king_inequality_2018}. Although a political solution with global cooperation is needed, technology can greatly aid climate action, and AI specifically holds a lot of potential for both mitigation and adaptation. \citet{rolnick_et_al_tackling_2019} outline the wide range of machine learning applications to climate action, from enhancing efficiency in transport and infrastructure, to advancing the energy transition by improving renewable energy technologies. Other examples include using ML models for more accurate weather and climate forecasts \cite{hwang_improving_2018} and applying deep learning to improve climate models \cite{rasp_deep_2018} or advance earth science more broadly \cite{maskey_earth_2018}. Meanwhile, the number of companies using AI to offer ‘climate services’ has surged for example through monitoring environmental risks (eg Ecometrica \cite{ecometrica_environmental_2018}), predicting extreme weather events (eg Jupiter \cite{jupiter_intelligence_jupiter_2019}), or providing data to assess general climate risks (eg Acclimatise \cite{acclimatise_building_2019}). Critics warn that the exclusive nature of these commercial services could exacerbate inequality, essentially enabling those who can afford it to protect themselves and even profit from climate change, while others suffer its impact \cite{dembicki_will_2019}. Similarly, a recent UN report warned of ‘climate apartheid’, where the impact of climate change mirrors existing lines of wealth and power \cite{noauthor_world_2019}. Consequently, and having noted that accountability for individual agents is insufficient, ensuring that AI fulfills its potential for good without exacerbating issues like climate inequality requires a structural approach to understand and mitigate systemic risks that could arise from AI’s interaction with the social, economic and political dimensions of climate change.

It is useful to consider a hypothetical example to elucidate what such systemic risks might look like, looking at food security as one climate-related issue where AI is well-positioned to help. A range of companies already use AI in relation to agriculture (eg \cite{indigo_ag_about_2018}) or specifically climate impact on agriculture (eg \cite{awhere_weather_2019}), but AI could also monitor food security in real time, as well as give longer-term warnings through ‘spatially localized crop yield predictions’ \cite{rolnick_et_al_tackling_2019}. The IPCC’s 'Climate Change and Land' report highlights that ensuring global food security requires understanding the impact of climate change \cite{arneth_et_al_ipcc_2019}. An AI-enabled model that predicts the impact of climate change on land and agricultural production would therefore be incredibly valuable. It would allow for pre-emptive climate action such as policies for more sustainable land use to mitigate yield loss caused by land degradation, while also anticipating short-term food shortages so that proactive policies can ensure continued access to food as opposed to emergency responses. However, it would interact with existing socio-economic and political structures shaping food production and distribution, since food security is equally a systemic issue requiring a political solution. The 2008 food crisis exemplified this, when technological solutions could not have prevented food insecurity. Although climate played a role in diminished yields, other factors were more important such as speculation driving up food prices, increased biofuel production, and diminishing buffer stocks and investment levels in agriculture \cite{mittal_2008_2009}. An AI platform predicting food supply would therefore need a structural perspective to understand the political and socio-economic factors that determine food security.

In our example several systemic risks can be hypothesized including: price hikes, hoarding of food supplies, environmentally unsustainable practices, conflict over arable land and increased inequality. Consider a scenario in which this AI platform reveals that in five years maize yields will drop by 20\% in region X which represents 10\% of global production, not wholly unlikely given the expected 7.4\% decline at one degree temperature rise \cite{zhao_et_al_temperature_2017}. As scarcity increases the value of goods, this could lead to a drastic price rise with devastating consequences for region X and cascade effects for the circa 820 million food insecure people worldwide who are most impacted by food price rises \cite{torrero_cullen_et_al_state_2019}. Additionally, maize growers might hoard their supplies, thus exacerbating price spikes and increasing the likelihood of food insecurity elsewhere as witnessed in 2008 when large maize-exporting countries imposed export bans \cite{tigchelaar_future_2018}. Since yield losses will differ between countries \cite{zhao_et_al_temperature_2017}, another scenario could be that a large country with considerable economic and military strength is found to suffer more from land degradation than its less powerful neighbouring country, providing an incentive to annex arable land. It could also lead to deforestation to make land available or depletion of other natural resources. In these scenarios, the AI platform would need to carefully consider how it shares its information, while engaging with social scientists to adopt a structural approach to understand and mitigate such risks. This could mean not sharing information freely with the highest bidder but only with appropriate actors and under specific conditions, while coordinating with regulatory bodies that monitor food prices or the use of natural resources for example. Moreover, it would have to engage with policymakers to ensure its benefits are not undermined by socio-economic and political factors.

\section{Conclusion}

Current literature on AI ethics addresses the need to minimise risks of AI systems through the prevention of accidents and misuse. This paper has shown that this understanding of AI risks needs to be complemented by a structural approach to account for systemic risks, in particular when AI systems engage with systemic issues such as climate change where technology’s impact is affected by the socio-economic and political context. Since there is no mathematical solution this paper cannot offer a technical road map, but we do suggest several ways our findings can be translated into practice. First, ethics codes must adopt both an agency and structural approach to encompass the wide range of AI risks. Such codes should become central to AI research and development, but are equally relevant to other stakeholders such as policy makers and users of AI systems. Moreover, since international cooperation is needed to regulate the ethical development and use of AI and prevent 'ethics shopping', ethics codes need to be standardised at an international level \cite{cihon_standards_2019}. The G20’s recommendations for ‘trustworthy AI’ are promising in calling for internationally comparable metrics and cooperation to ensure AI is beneficial ‘for people and the planet’, although it needs to be broadened to include systemic risks \cite{noauthor_g20_2019}. Second, interdisciplinary collaboration is needed in the formulation of ethics guidelines and their translation into policies, with an inclusive and democratic process to determine what constitutes 'AI for good'. Such collaboration between ‘governments, industry, academia and civil society’ \cite{gutteres_secretary-generals_2019} at a national and international level would also facilitate the implementation of policies and the creation of governance structures required to put abstract principles into practice. Finally, new regulations will require mechanisms to enforce them, such as independent regulators. At a national level this could involve legislation such as liability laws, something which the EU’s ‘expert group on liability and new technologies’ is currently investigating \cite{noauthor_liability_2019}. For companies this could mean creating an ethics committee and undertaking thorough, comprehensive risk assessments. Although enforcement at the global level remains the biggest challenge, the fact that systemic risks are also collective risks should provide a strong incentive for international cooperation \cite{zwetsloot_thinking_2019}. Lessons can be drawn from other sectors, for instance international regulation on nuclear weapons or preventing systemic risks in the financial sector. The example from the medical domain mentioned in section \ref{systemic risks} suggests an international code of publication ethics where editors are responsible for the wider ethical implications of publishing findings \cite{ploug_should_2018}. Research is needed to answer similar questions for AI around how ethics codes should be implemented and by whom. Presumably democratic governments are best positioned to lead this as they have the capacity to enforce norms and in theory would prioritize the common good rather than commercial profit or academic status. Nevertheless, AI ethics remains deeply contentious as it 'is effectively a microcosm of the political and ethical challenges faced in society' \cite{mittelstadt_ai_2019}. As the specifics of AI governance are beyond the scope of this paper, these are merely preliminary suggestions but we hope that demonstrating the need for a structural approach to complement the current AI ethics discourse will encourage further research on systemic risks, how they can be integrated into policies and how these are then best enforced to ensure ethical AI.

\medskip

\small

\bibliographystyle{plainnat}

\bibliography{neurips_2019}

\end{document}